\documentclass[sigconf]{acmart}
\renewcommand\footnotetextcopyrightpermission[1]{} 
\usepackage{multicol}
\usepackage{lipsum}

\AtBeginDocument{%
  \providecommand\BibTeX{{%
    \normalfont B\kern-0.5em{\scshape i\kern-0.25em b}\kern-0.8em\TeX}}}
\usepackage{graphicx}
\usepackage{subcaption}
\usepackage{amsmath}
\usepackage{amsmath}
\usepackage{bm}
\usepackage{amssymb}

\begin{document}

\title{DragEntity:Trajectory Guided Video Generation using Entity and Positional Relationships}


\author{Zhang Wan}
\affiliation{%
  \institution{Institute of Computing Technology, Chinese Academy of Sciences University of Chinese Academy of
Sciences}
  \city{Beijing}
  \country{China}}
\email{wanzhang22b@ict.ac.cn}
\orcid{0009-0006-0930-9284}

\author{Sheng Tang}
\authornote{
Corresponding author: Sheng Tang. The authors are also with Key Lab of Intelligent Information Processing, Institute of
Computing Technology, Chinese Academy of Sciences.}
\affiliation{%
  \institution{Institute of Computing Technology, Chinese Academy of Sciences University of Chinese Academy of Sciences}
  \city{Beijing}
  \country{China}}
\email{ts@ict.ac.cn}
\orcid{0000-0003-3573-2407}

\author{Jiawei Wei}
\affiliation{%
  \institution{Zhengzhou University}
  \city{Zhengzhou}
  \country{China}}
\email{weijiawei@gs.zzu.edu.cn}
\orcid{0009-0004-2154-9913}

\author{Ruize Zhang}
\affiliation{%
        \institution{Institute of Computing Technology, Chinese Academy of Sciences University of Chinese Academy of Sciences}
        \city{Beijing}
        \country{China}
}
\email{zhangruize21b@ict.ac.cn}
\orcid{0000-0001-5999-2866}

\author{Juan Cao}
\affiliation{%
    \institution{Institute of Computing Technology, Chinese Academy of Sciences University of Chinese Academy of
Sciences}
  \city{Beijing}
  \country{China}}
\email{caojuan@ict.ac.cn}
\orcid{0000-0002-7857-1546}



\begin{abstract}
In recent years, diffusion models have achieved tremendous success in the field of video generation, with controllable video generation receiving significant attention. However, existing control methods still face two limitations: Firstly, control conditions (such as depth maps, 3D Mesh) are difficult for ordinary users to obtain directly. Secondly, it's challenging to drive multiple objects through complex motions with multiple trajectories simultaneously. In this paper, we introduce DragEntity, a video generation model that utilizes entity representation for controlling the motion of multiple objects. Compared to previous methods, DragEntity offers two main advantages: 1) Our method is more user-friendly for interaction because it allows users to drag entities within the image rather than individual pixels. 2) We use entity representation to represent any object in the image, and multiple objects can maintain relative spatial relationships. Therefore, we allow multiple trajectories to control multiple objects in the image with different levels of complexity simultaneously. Our experiments validate the effectiveness of DragEntity, demonstrating its excellent performance in fine-grained control in video generation.
\end{abstract}




\maketitle

\section{Introduction}
The task of video generation, such as the popular Text-to-Video (T2V) generation task \cite{blattmann2023align,
chen2023videocrafter1,he2022latent}, aims to create various high-quality videos based on given text prompts. Unlike image generation \cite{rombach2022high,saharia2022photorealistic}, which concentrates on producing a single image, video generation involves crafting a sequence of images that exhibit consistent and smooth motion. Consequently, motion control is pivotal in video generation and has garnered substantial attention in recent research on various control methods.

\begin{figure}[t]
\centering
\includegraphics[width=3.5in]{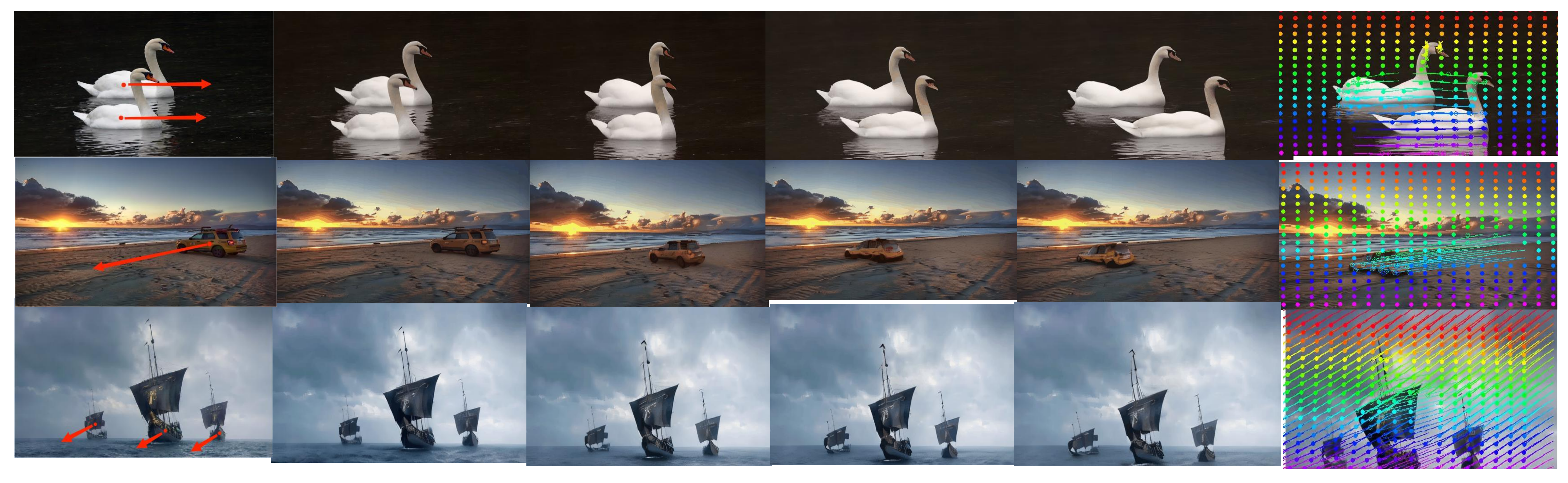}
\vspace{-0.5cm}
\caption{Some examples. Input an image and a trajectory, and output a video. We select and display the 1st, 5th, 15th, and 20th frames for demonstration.}
\label{fig1}
\vspace{-1em} 
\end{figure}

In the domain of controllable video generation, previous research has mainly concentrated on image-to-video generation. These methods \cite{lotter2016deep,srivastava2015unsupervised, chiappa2017recurrent} utilize an initial frame image as the control condition to generate video sequences. However, using an image as the initial frame to control the conditions can result in difficulty managing the content of subsequent video frames, making it challenging to generate the video that the user desires. Consequently, some research \cite{singer2022make,hong2022cogvideo} has shifted towards the text-to-video domain, using long texts or prompt words to constrain the semantic content. Nevertheless, due to the inherent ambiguity and subjectivity of language, the information available as control conditions remains limited, especially in terms of complex movements of objects.

In the field of video generation, trajectory-based control has emerged as a user-friendly method, attracting increasing attention from researchers. CVG \cite{hao2018controllable} and C2M \cite{ardino2021click} encode images and trajectories by predicting optical flow maps and performing feature warping operations to achieve controllable video generation. However, warping operations often lead to unnatural distortions. To address this issue, II2VB \cite{blattmann2021understanding} and iPOKE \cite{blattmann2021ipoke} condition on an initial frame and a local poke, allowing for the sampling of object kinematics and establishing a one-to-one correspondence with the resulting plausible videos. Similarly, MCDiff  \cite{wang2024videocomposer} predicted future frames in an autoregressive manner through diffusion delay. While MCDiff relies on HRNet \cite{wang2020deep} to extract 17 key points for each individual, it can only control the movements of humans. Additionally, the generated videos do not move along the prescribed trajectory paths. Furthermore, MCDiff overlooks the generation of open-domain videos, significantly limiting its practical application value.

One representative work, DragNUWA \cite{yin2023dragnuwa} , encodes sparse trajectories into dense flow space sequences and then uses these sequences as supervisory signals to control object motion. Similarly, MotionCtrl \cite{wang2023motionctrl} presents a unified and flexible motion controller that encodes the movement paths of objects into a vector field, utilizing this field along with camera poses and object trajectories to control the motion of generated videos. 

\begin{figure*}[t]
\centering
\includegraphics[width=1.0\linewidth]{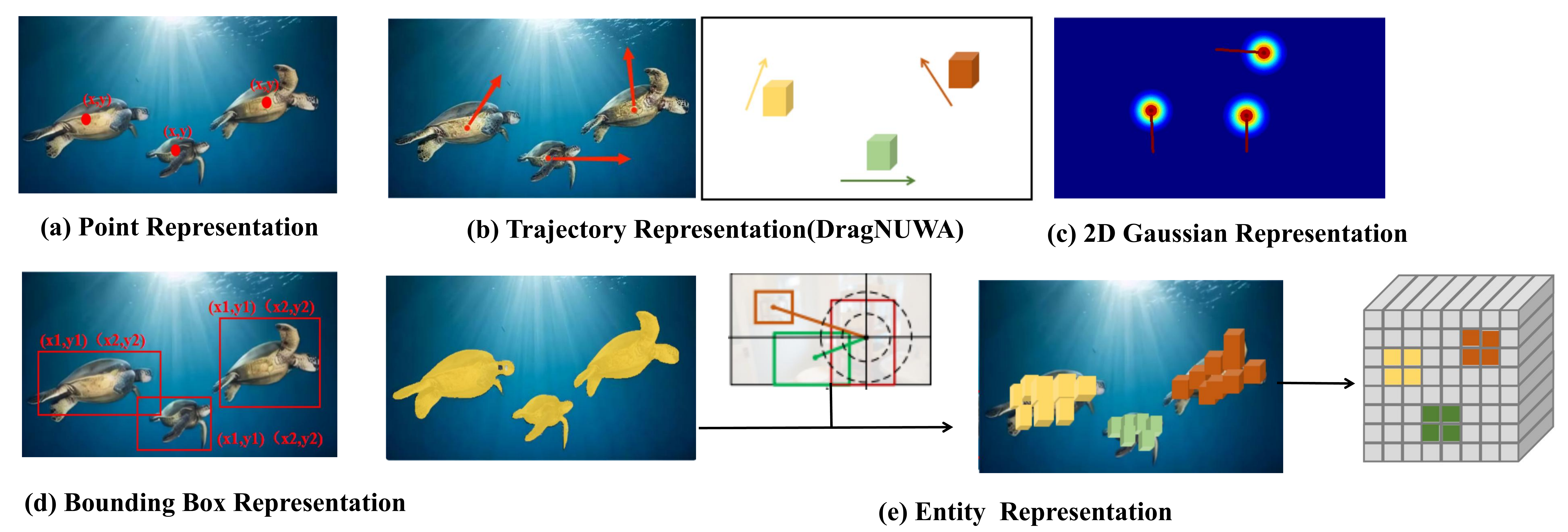}
\vspace{-0.5cm}
\caption{ Comparison of different representation modeling methods:
(a) Point Representation: Represents an entity using coordinate points (x, y).
(b) Trajectory Graph: Represents the trajectory of an entity using a trajectory vector graph.
(c) 2D Gaussian Distribution: Represents an entity using a two-dimensional Gaussian mapping.
(d) Box Representation: Represents an entity using a bounding box.
(e) Entity Representation: Represents an entity using latent features that include spatial relationships between objects. }
\label{fig2}
\end{figure*}

It should be noted that the above-mentioned research still has some issues. Firstly, it is evident that a single pixel cannot represent an entity, and dragging a single pixel cannot accurately control the corresponding entity. Secondly, when dealing with human datasets such as TED talks and Human 3.6M, multiple complex trajectories can cause inconsistent pixel movement in different dragged areas, leading to severe distortion of the human body. In fact, solving this problem requires clarifying two conditions: 1) What is the entity? Identify the specific area or object to be dragged. 2) How to drag? Separate the dragged area from the background, and ensure that the relative spatial positions between multiple dragged areas remain consistent. To address the first challenge, we use interactive segmentation methods \cite{kirillov2023segment,wang2023seggpt} to select entities. For instance, using SAM \cite{kirillov2023segment} in the initial frame allows us to conven on a given image allows us to freely choose the region we want to control.
To achieve accurate individual human twisting, it is necessary to employ fine-grained segmentation, treating limbs and the head as different entities. In contrast, the second technical challenge poses a greater difficulty. To address this issue, we proposes a novel entity representation method that integrates the positional relationships between entities to achieve precise motion control of objects in videos, as shown in Figure \ref{fig1}.

Some works \cite{chen2023anydoor,gu2023videoswap,tang2023emergent} have shown the effectiveness of feature representation at the entity level in images. Anydoor \cite{chen2023anydoor}  utilizes the capabilities of Dinov2 \cite{oquab2023dinov2}  for object customization, while VideoSwap \cite{gu2023videoswap} and DIFT  \cite{tang2023emergent} utilize the capabilities of diffusion models \cite{rombach2022high} for video editing tasks based on entity features. DragAnything \cite{wu2024draganything} integrates instance segmentation techniques to handle video generation tasks based on diffusion models. Inspired by these works, we propose DragEntity, which employs the latent features to represent individual entities. For the human body, this representation is further refined to encompass 12 entity parts, including the torso and limbs. Furthermore, we believe that the relative spatial relationship between entities is crucial for modeling the motion of objects. For example, a swan won't swim into the lake, and a person's head cannot be lower than their shoulders.

In our work, we utilize SAM\cite{kirillov2023segment} as the base segmentation model (for human bodies, LIP\cite{gong2017look}  is used for fine-grained segmentation). In contrast to previous work, we design an entity representation mechanism. First, the segmentation model is used to obtain entity masks in the initial frame. Then, based on the mask of each entity, the central coordinates are extracted to generate the entity representation. Finally, relative spatial position information is integrated into the entity representation through position relationships awareness module. 

Our main contributions are summarized as follows:
\begin{itemize}
    \item Unlike the paradigm of dragging pixels, we propose a method for dragging entities that enables true entity-level motion control and representation, ensuring the structural integrity of the entity during the dragging process.
\end{itemize}
\begin{itemize}
    \item We introduce modeling of the relative spatial positions between objects to prevent the generation of highly unrealistic motion videos caused by trajectory dragging.
\end{itemize}
\begin{itemize}
\item We have carried out experiments to confirm the efficacy of DragEntity, showcasing its excellent performance in precise control of video generation.
\end{itemize}

\begin{figure*}[htb]
\centering
\subfloat[]{\includegraphics[width=3.5in,height=1.95in]{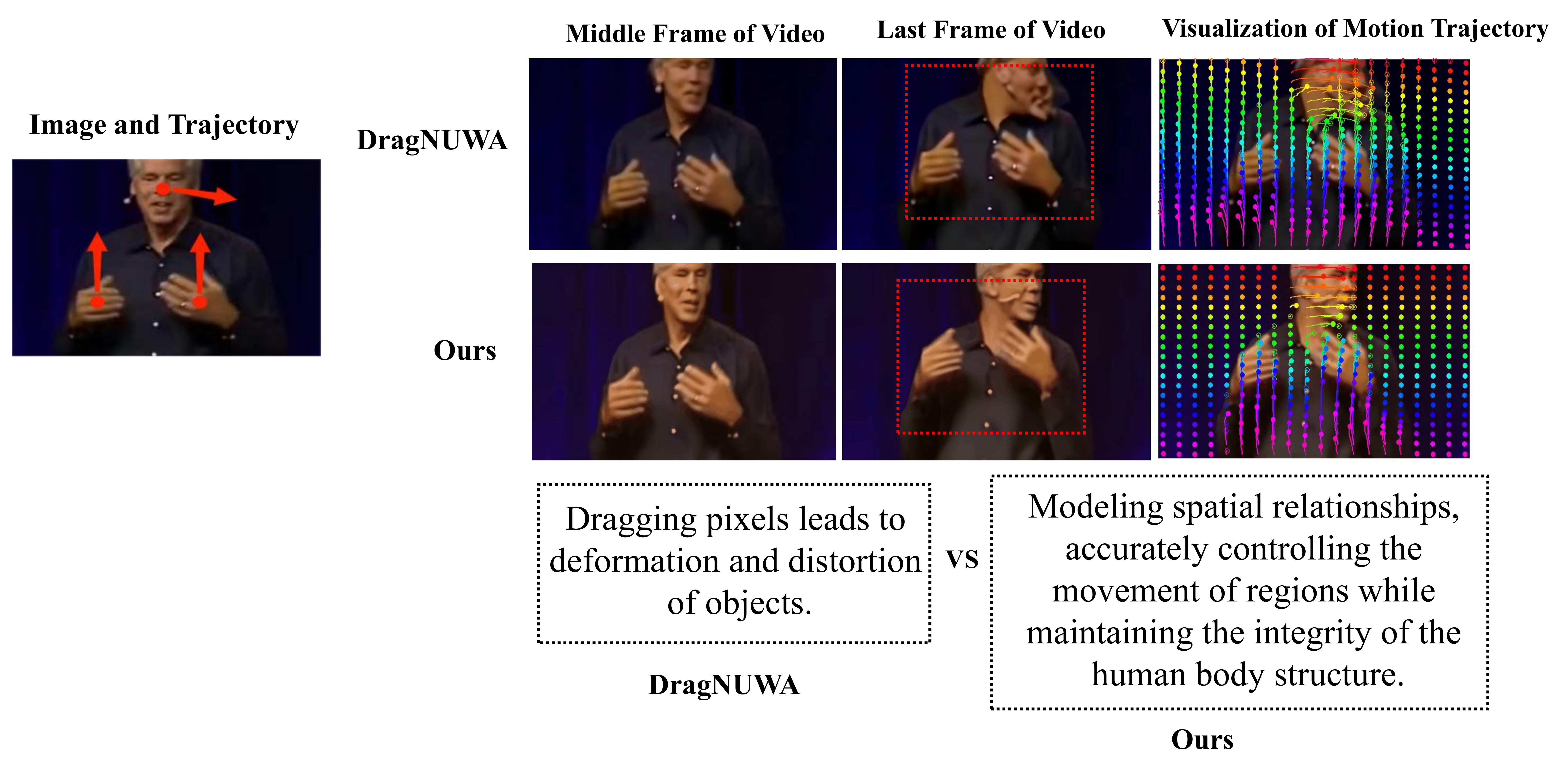}%
}
\hfil
\subfloat[]{\includegraphics[width=3.5in,height=2.0in]{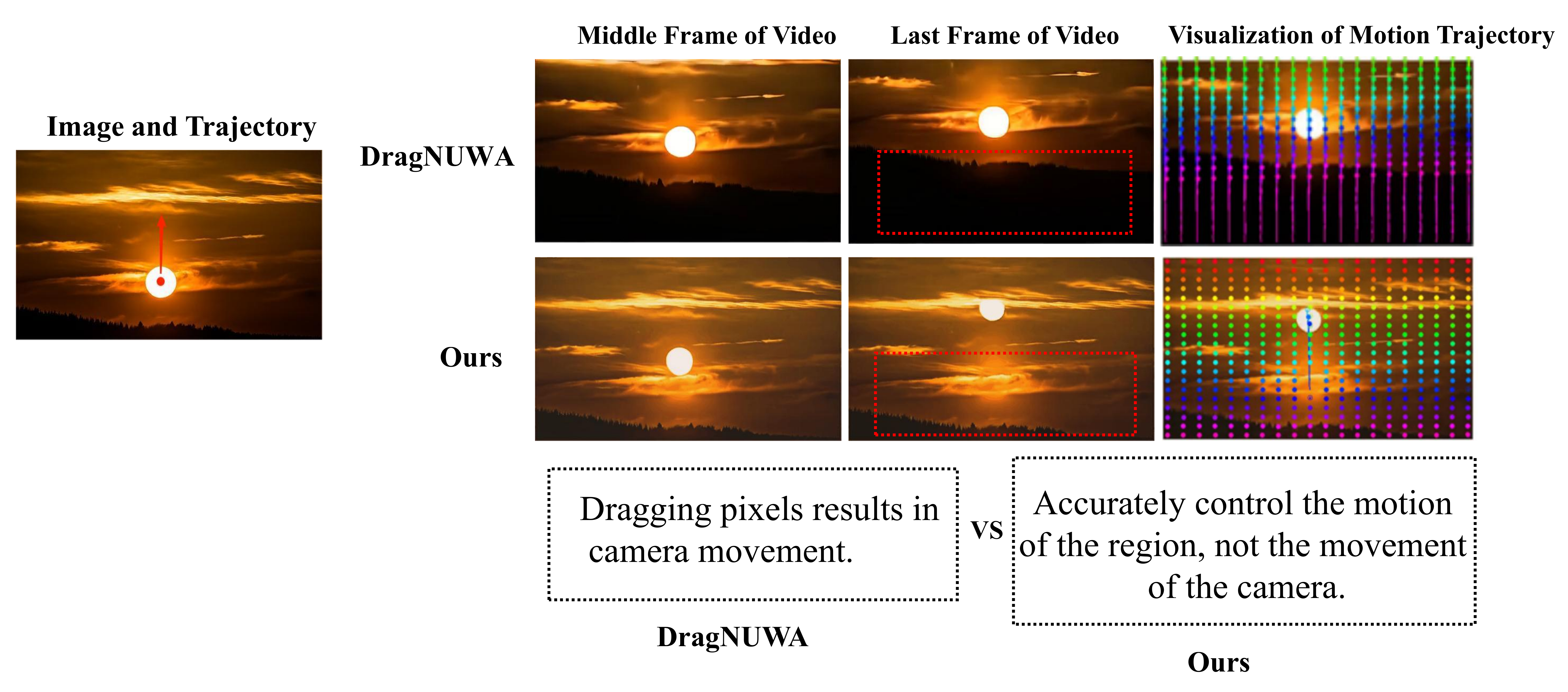}}%
\label{fig_second_case}
\caption{
Experiments on the motivation for entity representation. Existing methods (DragNUWA and MotionCtrl ) involve directly dragging pixels, which cannot precisely control the target, leading to camera motion or target structure distortion. In contrast, our method utilizes entity representation and models spatial relative positions to achieve accurate control.}
\label{fig3}
\end{figure*}

\section{Related Work}
\subsection{ Image and Text Guided Video Generation}
In recent years, the field of text-to-video generation has garnered substantial attention \cite{wu2021godiva,wu2022nuwa,hong2022cogvideo,singer2022make}, using textual descriptions to guide the semantic creation of video content. However, text alone often falls short in conveying precise spatial information. To address this, MAGE \cite{wu2022nuwa}  employs a text-image-to-video strategy, merging the semantic elements of text with spatial information from images to achieve detailed video control. Likewise, GEN-1  \cite{rombach2022high}  utilizes cross-attention mechanisms to combine depth maps and text for improved control. For extended video generation, text-image-to-video methods are also prominent. Phenaki \cite{villegas2022phenaki}, for example, adopts an autoregressive approach, using previous frames and text to generate subsequent ones, facilitating long video production. NUWA-XL \cite{yin2023nuwa}  uses a hierarchical diffusion model to iteratively fill in intermediate frames based on earlier frames and text. I2VGEN-xl \cite{zhang2023i2vgen} introduces a cascaded network that enhances performance by decoupling semantic and spatial components, ensuring alignment with static images as a key reference. In addition to academic research, the industry has also made notable contributions, such as Gen-2\cite{esser2023structure} and SORA \cite{brooksvideo}. Despite these advancements, the area of controllable video generation still holds significant potential for further development.
\subsection{Trajectory-based Video Generation}
Early trajectory-based methods\cite{blattmann2021ipoke,blattmann2021understanding}  often used optical flow or recurrent neural networks for motion control. TrailBlazer \cite{ma2023trailblazer} enhances video synthesis controllability by utilizing bounding boxes to guide subject movement. DragNUWA \cite{yin2023dragnuwa} converts sparse strokes into a dense flow space to control object motion. Similarly, MotionCtrl \cite{wang2023motionctrl} encodes each object’s trajectory coordinates into a vector map for motion guidance. These approaches can be divided into two categories: Trajectory Map (point) and Box Representation, as shown in Figure \ref{fig2}. Box representation methods, such as TrailBlazer, are restricted to instance-level objects and cannot handle backgrounds like starry skies. Current Trajectory Map methods, including DragNUWA and MotionCtrl, are relatively basic as they ignore the semantic aspects of entities, making a single point inadequate to represent an entity. Although DragAnything incorporated entity representation, it neglected positional relationships, leading to subpar performance when managing multiple trajectories. In this paper, we present DragEntity, which employs entity representation along with positional relationship data for trajectory guidance, maintaining structural integrity and spatial coherence during trajectory manipulation.
\begin{figure*}[t]
\centering
\includegraphics[width=1.0\linewidth]{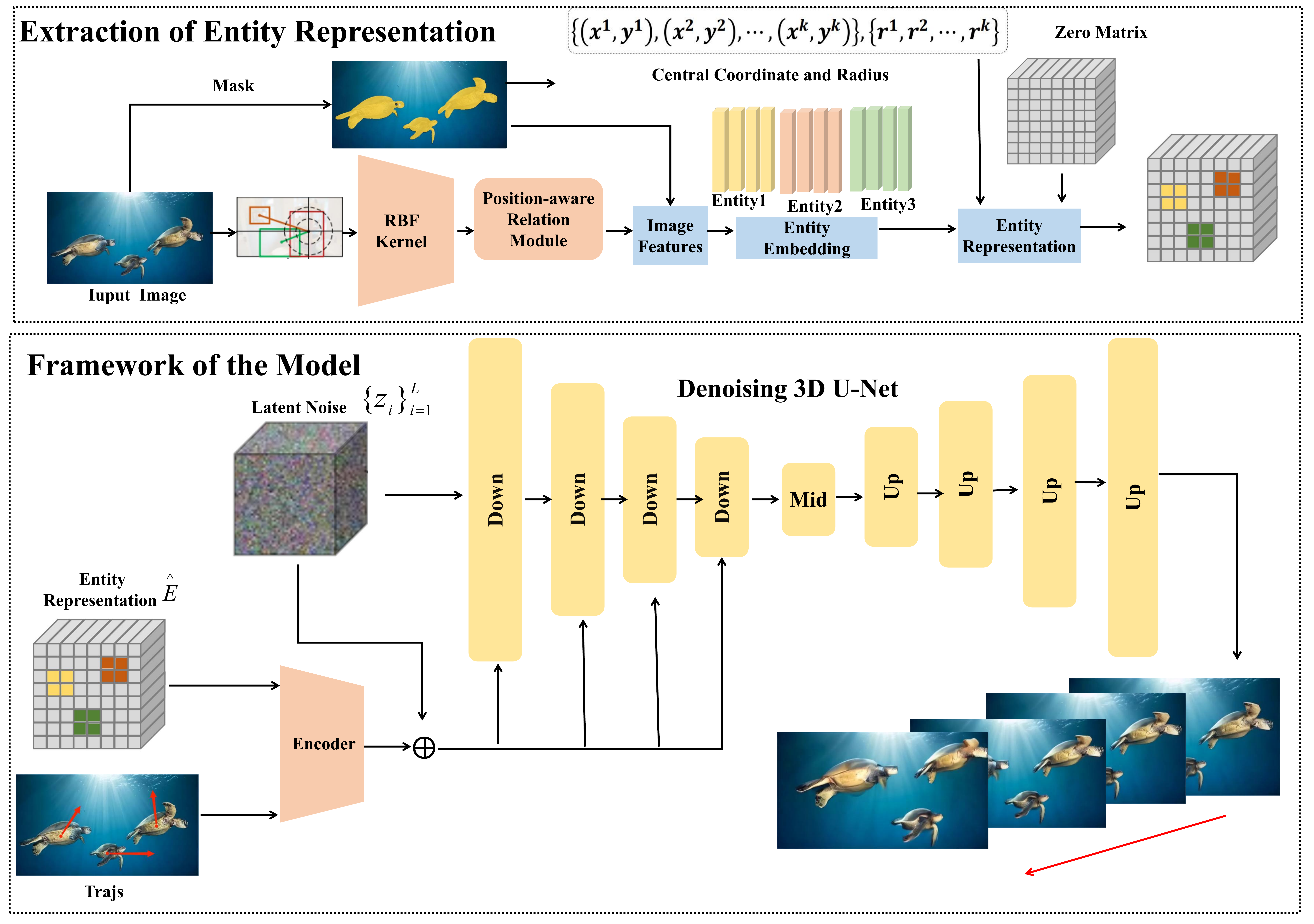}
\vspace{-0.5cm}
\caption{ Model Framework. This image consists of two parts:
(a) Entity Semantic Representation Extraction. Latent features are extracted based on entity mask indices, integrating the relative spatial relationships between objects to form their respective entity representations.
(b) Main Framework. Based on the SVD\cite{blattmann2023stable}  model, it utilizes the corresponding entity representations to precisely control motion. }
\label{fig4}
\end{figure*}
\section{Method}
\subsection{Task Definition and Motivation}
{\bf Task Definition.} 
The task of trajectory-based controllable video generation requires the model to generate videos based on a given image and motion trajectory. Given a point trajectory $(x_1, y_1), (x_2, y_2)$, $\dots$, $(x_L, y_L)$, where $L$ represents the length of the video, the conditional denoising autoencoder $\epsilon_{\theta}(z, c)$ is utilized to generate videos corresponding to the motion trajectory. In this paper, the guiding signal $c$ contains three types of information: trajectory points, the first frame of the video and entity masks of the initial frame.

{\bf Motivation.}  
Recently, several trajectory control approaches, such as DragNUWA  \cite{yin2023dragnuwa} and MotionCtrl \cite{wang2023motionctrl}, have utilized trajectory points to manage object motion. These techniques typically alter the corresponding pixels or pixel regions directly based on the given trajectory coordinates or their derivatives. However, they neglect a crucial issue: as illustrated in Figure \ref{fig3}, the pixels or pixel regions directly manipulated by the trajectory do not necessarily correspond to the entity intended for control. Consequently, dragging these points does not enable object motion based on trajectory control. As illustrated in Figure \ref{fig3}, we visualize the trajectory changes of each pixel in the generated video based on the co-tracker \cite{karaev2023cotracker}. We can observe that:

(1) It is apparent that a single pixel or a small group of pixels on an object cannot fully represent the entire entity (Figure \ref{fig3} (b)). From the pixel motion trajectory observed in DragNUWA, it is evident that dragging a pixel of the sun does not result in the sun moving; instead, it causes the camera to shift upwards. This clearly illustrates that a single pixel or a few pixels cannot represent the entirety of the sun, preventing the model from comprehending the true intention behind the trajectory. We employ an entity representation of objects as a more direct and effective method to accurately control the area we manipulate (the selected area), while keeping the rest of the image unchanged.

(2) As shown in Figure \ref{fig3} (a), when multiple trajectories influence the same object, the motion of each part must maintain relative spatial relationships to preserve the object's structural integrity. In contrast, in videos generated by DragNUWA, we observed that different parts of the human body move independently under the control of trajectories, leading to abnormal distortions. However, what we expect is for the human body to perform various combined bodily movements as a whole under the guidance of multiple trajectories, rather than each body part acting separately.

Based on these new insights and observations, we propose a new entity representation that integrates the spatial relationships of objects, extracting the latent features of the objects we want to control for their representation. As shown in Figure \ref{fig3}, the visualization of motion trajectories indicates that our method can achieve more precise motion control. For example, in Figure \ref{fig3}(a), our method can accurately control the combined movements of turning the head and raising the hand, whereas DragNUWA only drags the corresponding pixel areas for motion, without considering the relationships between different parts, leading to abnormal appearance distortions. In Figure \ref{fig3}(b), our method can accurately control the rising of the sun, whereas DragNUWA interprets it as a camera displacement.

Based on the SVD \cite{blattmann2023stable} model, the architecture mainly comprises three components: a denoising diffusion model (3D U-Net \cite{ronneberger2015u}) that learns the denoising process across spatial and temporal dimensions, an encoder and decoder that translate the supervisory signal into the latent space and reconstruct the denoised latent features back into video form, as illustrated in Figure \ref{fig4}. Drawing inspiration from ControlNet \cite{zhang2023adding}, we employ a 3D Unet to encode our guiding signal and subsequently apply it to the decoder block of the SVD's denoising 3D Unet. Unlike previous approaches, we have developed an entity representation mechanism that incorporates the relative spatial relationships of objects, allowing for trajectory-based controllable video generation.

\subsection{Entity  Representation includes Spatial Relationships}
The conditional signal of our method requires corresponding entity representations. In this section, we will describe how to extract these representations from the first frame of the image.

\begin{figure}[t]
\centering
\includegraphics[width=0.9\linewidth]{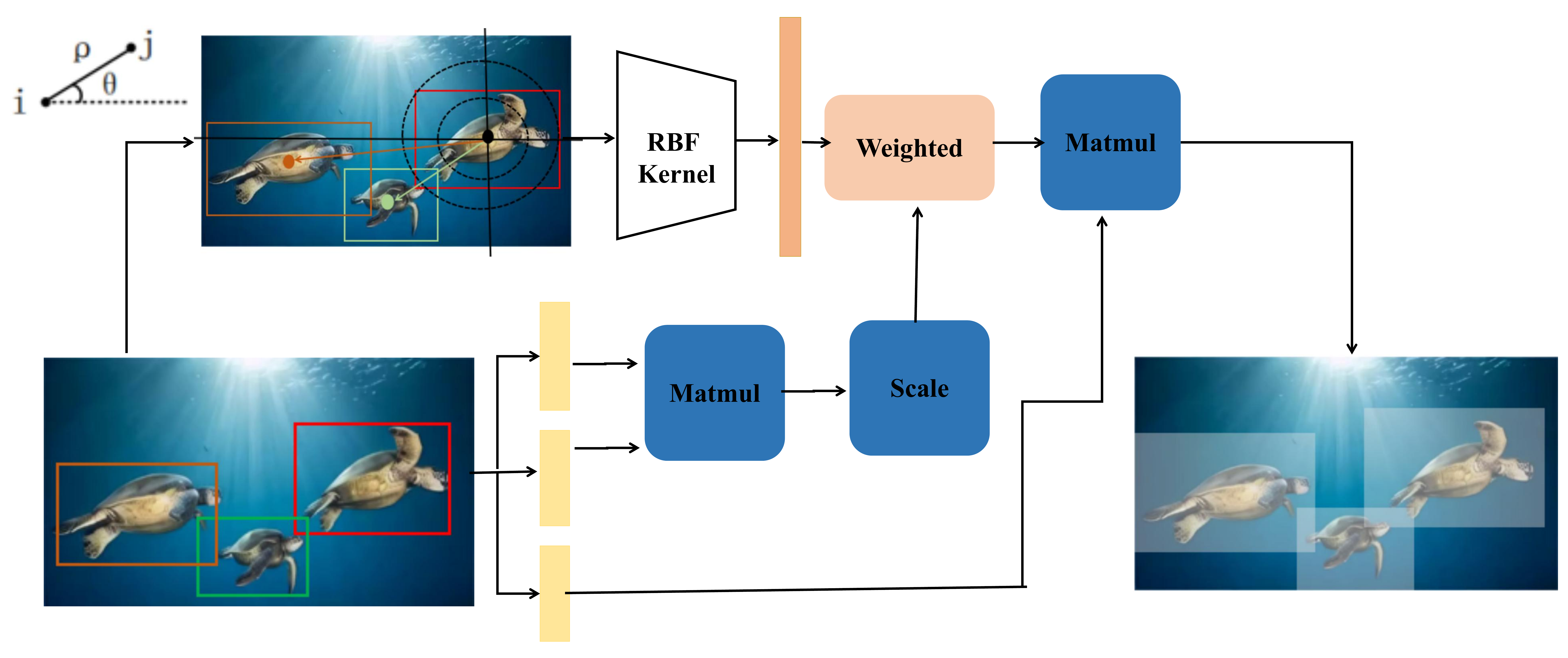}
\caption{ Image Position-Aware Relationship Module. The entity representation includes more information about the relative spatial relationships between objects. }
\label{fig5}
\end{figure}

{\bf  Position-aware relation.} Inspired by ParNet\cite{xia2019parnet}, we propose a position-aware relationship module that simultaneously captures semantic and spatial object-level relationships. It is designed to enrich the representation of objects by adaptively focusing on the spatially relevant and semantically relevant parts of the input image, as shown in Figure \ref{fig5}. 

In the spatial relationship branch, for each object \(i\), a polar coordinate system is centered at \(i\), and the four-dimensional bounding box position \(p_j = (x, y, w, h)\) of object \(j\) is transformed into a polar coordinate vector \((\rho_j, \theta_j)\). This is because representing the spatial direction between the centers of bounding boxes \(i\) and \(j\) is very effective. We believe that in describing the positional relationships between objects, absolute positions are rarely used. Instead, relative positions are widely utilized (for example, the head is directly above both arms). Therefore, our method focuses on the relative positions of objects rather than their absolute positions.

Low-dimensional relative positions are embedded into higher dimensions using a set of Gaussian kernels with learnable means and covariances. This allows the spatial relationship between objects \(i\) and \(j\) to be easily distinguishable. Experimentally, the dimension of the spatial relation after embedding is set to 
\(d_p = 64\). The kernel operator for an object \(j\) centered at \(i\) is defined as follows:

\begin{align}
\omega_{\rho_j} &= \exp\left(-\frac{\|\rho_j - \rho_0\|^2}{2\sigma_{\rho}^2}\right) \tag{1} \\
\omega_{\theta_j} &= \exp\left(-\frac{\|\theta_j - \theta_0\|^2}{2\sigma_{\theta}^2}\right)\tag{2}
\end{align}
Where \(\rho_0\) and \(\sigma_{\rho}\) are the learnable means and covariances of Gaussian distributions for relevant distance, and \(\theta_0\) and \(\sigma_{\theta}\) are the learnable means and covariances of Gaussian distributions for relevant angle. 

We combine the relative distance and angular relationship between objects \(i\) and \(j\)  using a scaling function. This allows the strength of spatial relationships between objects to be weighted based on spatial orientation. The combined spatial weight for the image is represented as \(\omega_p\):

\begin{equation}
\omega_p = \frac{ \omega_{\rho_j} \omega_{\theta_j}}{\sum_{j=1}^N \omega_{\rho_j} \omega_{\theta_j}} \tag{3}
\end{equation}

where \(N\) is the number of objects in the image.

In another branch, image features \( V \) is linearly transformed by \( f(\cdot) \). Semantic relation \( \omega_s \) is computed as in Eq.~(4). Dot-product attention is employed in our algorithm with a scaling factor \( \frac{1}{\sqrt{d_v}} \), where \( d_v \) is the dimension of object features \( v \).
\begin{equation}
\omega_s = \frac{f(V)^T f(V)}{\sqrt{d_v}} \tag{4}
\end{equation}
where \( d_v \) is the dimension of object feature \( V \).

The intra-image relation weight \( \omega_I \) indicates both the semantic and spatial impact from object \( j \). Spatial relationship \( \omega_p \) of different objects is fused with semantic relationship \( \omega_s \) between objects through Eq.~(5). It is scaled in the range (0, 1) and can be regarded as a variant of softmax. \( \omega_I \) is computed as follows:
\begin{equation}
\omega_I = \frac{\omega_p \exp(\omega_s)}{\sum_{k=1}^N \omega_p \exp(\omega_s)} \tag{5}
\end{equation}
To adapt to flexible relationships, multi-head attention \cite{vaswani2017attention} is utilized, as different heads can concentrate on various aspects of the relationships. The multiple relational features from these heads are then aggregated in the following manner:
\begin{equation}
V_r = f(V) + \text{Concat}[(\omega_I f(V))_1, (\omega_I f(V))_2, \ldots, (\omega_I f(V))_K] \tag{6}
\end{equation}
\( K \) is the number of relation heads, which is typically set to be 6, same as Transformer.

{\bf  Entity Representation.} Using the image features \( V_r \), the corresponding entity embeddings can be obtained by indexing the coordinates from the segmentation mask. For convenience, average pooling is used to process the corresponding entity embeddings, resulting in the final embeddings \(\{e_1, e_2, \ldots, e_k\}\), where \( k \) represents the number of entities, and the channel size of each entity is \( C \).

To link these entity embeddings with the associated trajectory points, we start by initializing a zero matrix \( E \in \mathbb{R}^{H \times W \times C} \). We then embed the entity representations according to the trajectory sequence points, as illustrated in Figure\ref{fig4}. During training, the entity mask from the initial frame is used to extract the center coordinates \(\{(x_1, y_1), (x_2, y_2), \ldots, (x_k, y_k)\}\) of the entity, which serve as the starting points for each trajectory sequence. Using these center coordinate indices, we create the final entity representation \( \hat{E} \) by embedding the entity representations into the zero matrix \( E \).

With the entity's center coordinates \(\{(x_1, y_1), (x_2, y_2), \ldots, (x_k, y_k)\}\) from the first frame, we utilize Co-Tracker to track these points and generate the corresponding motion trajectories $\{ (x_{ki}, y_{ik}) \}_{i=1}^L $, where  \( L \) denotes the length of the video. This allows us to obtain the corresponding entity representation \(\{\hat{E}_i\}_{i=1}^L\) for each frame.

{\bf  Encoder for Entity Representation.} This encoder uses four convolutional blocks to handle the input guidance signal. Each block includes two convolutional layers and a SiLU activation function. After encoding, we combine the latent feature representations of the entity with the corresponding latent noise at each step.

\begin{equation}
\{\bm{\mathrm{R}_i}\}_{i=1}^L = \text{encoder}(\{\bm{\mathrm{\hat{E}}_i}\}_{i=1}^L) + \{\bm{\mathrm{Z}_i}\}_{i=1}^L \tag{7} \label{equ2}
\end{equation}
where \( \bm{\mathrm{Z}_i} \) represents the latent noise for the \(i\)-th frame. Subsequently, the feature set \(\{\bm{\mathrm{R}_i}\}_{i=1}^L\) is fed into the encoder of the denoising 3D Unet. This process yields four features at varying resolutions, which function as latent conditioning signals.

\subsection{ Training and Inference} 
In the training phase, it is essential to generate trajectories for Entity Representation. First, for each entity, we calculate its incircle based on the mask to find its center coordinates \((x, y)\) and radius \(r\). Using Co-Tracker, we then determine the trajectory of these center points \(\{(x_i, y_i)\}_{i=1}^L\), which represents the entity's motion path. For the entity representation, we embed the entity representation within the circle centered at 
\((x, y)\) with radius \(r\). 

\textbf{Loss Function.} We optimize the model using mean squared error (MSE). Given the corresponding entity representation \(\hat{E}\), the objective can be simplified to:
\begin{equation}
\mathcal{L}_\theta = \sum_{i=1}^{L} \bm{\mathrm{M}} \left\| \epsilon - \epsilon_\theta \left(\bm{x}_{t,i} \operatorname{cond}(\bm{\mathrm{\hat{E}}}_i) \right) \right\|_2^2 \tag{8}
\end{equation}
where \(E_\theta\) denotes the encoder for entity representations. 
The mask \(M\) is used to identify the entities in each frame of the images. The goal of optimizing our model is to accurately control the motion of the target object while preserving the generation quality of other objects and the background. Therefore, we apply the mask \(M\) to limit the MSE loss, ensuring that backpropagation affects only the specific regions we aim to optimize. By doing this, we can selectively influence only the desired areas during the training process, maintaining the integrity of the rest of the image.

\textbf{Inference of User-Trajectory Interaction.} 
During the inference process, users simply need to click to select the area they want to control with SAM \cite{kirillov2023segment} . For the human body, this is further refined into segmented areas for each part. Then, by dragging any pixel within that area, they form a reasonable trajectory. Our model can then accurately control the area based on this trajectory to generate a video corresponding to the desired motion.

\section{ Experiments}
\subsection{Experiment Settings}
\textbf{Implementation Details.} All our training is based on Stable Video Diffusion (SVD) \cite{blattmann2023stable}. Our experiments are conducted on PyTorch, utilizing Tesla A100 GPUs, with Adam as the optimizer. We perform a total of 100k training steps, with the learning rate set to 1e-6.

\textbf{Evaluation Metrics.} We employ two types of evaluation metrics: 1) Video Quality Assessment: We use Frechet Inception Distance (FID) and Frechet Video Distance (FVD) \cite{unterthiner2018towards} to evaluate image quality and temporal consistency. 2) Object Motion Control Performance: We assess the trajectory control capability using the Euclidean distance (ObjMC) between the predicted object trajectories and the ground truth trajectories.

\begin{figure*}[t]
\centering
\includegraphics[width=1.0\linewidth]{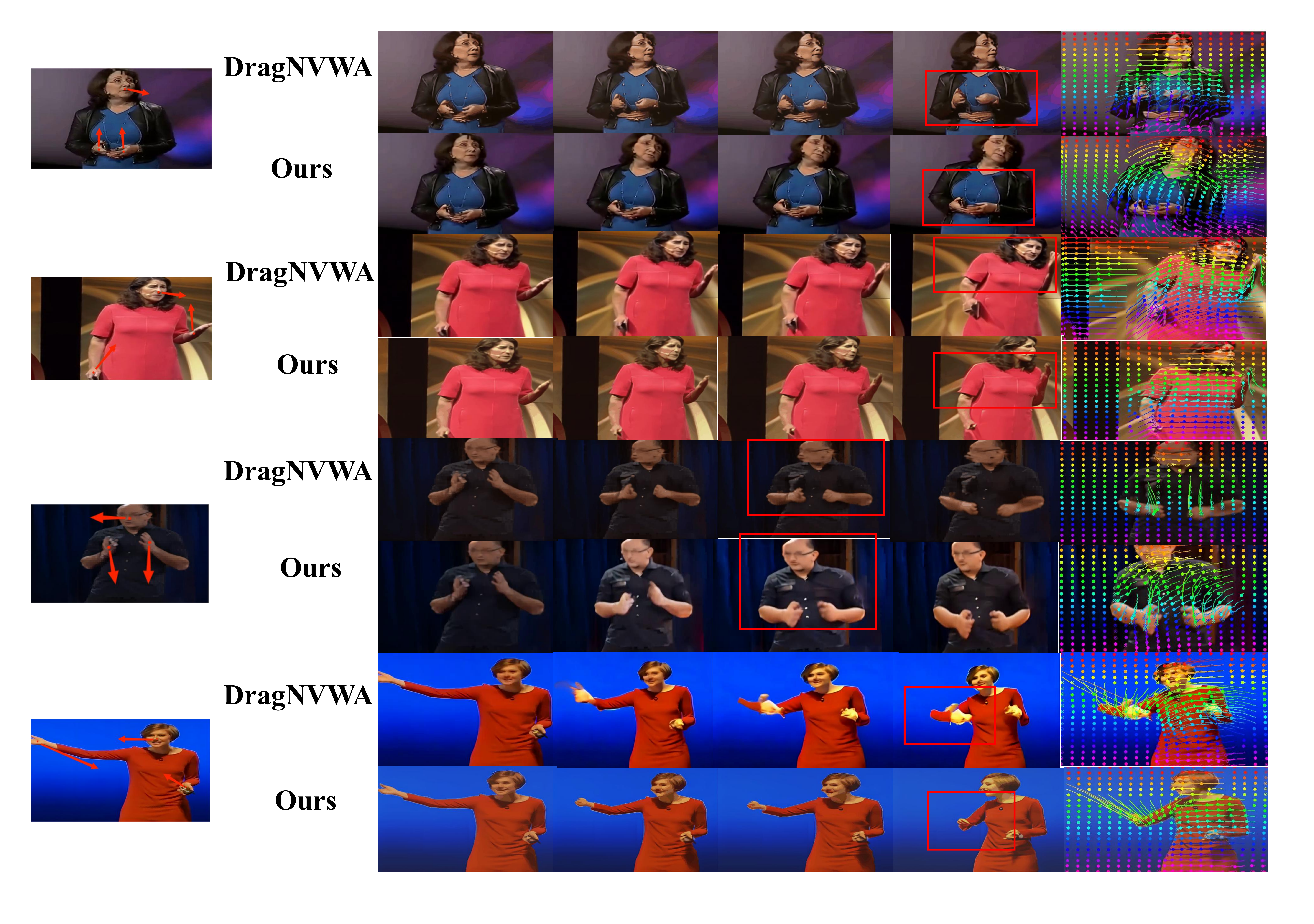}
\vspace{-0.5cm}
\caption{ Visual comparison on the TedTalk dataset. It can be observed that when multiple trajectories are active simultaneously on the human body, the DragNVWA model exhibits phenomena such as character distortion (third and seventh rows), artifacts (fifth row), and multiple hands (first row). Our model, while maintaining the basic skeleton of the human body, is able to move accurately according to the trajectories.}
\label{fig6}
\end{figure*}

\textbf{Datasets.} We use the VIPSeg\cite{miao2022large} and WebVid\cite{oquab2023dinov2} as our training sets, and employed a collaborative tracker to obtain the corresponding motion trajectories, which served as annotations. We adopt  WebVid\cite{oquab2023dinov2} as our testsets.
\subsection{Comparative Experiments}
\textbf{Comparisons with State-of-the-Art Methods.} Video Quality Evaluation on on the WebVid datasets. Table \ref{tab1} presents a comparison of our method with the existing SOTA methods on the WebVid dataset. It can be seen that in terms of generated image quality (FID), our score reached 34.5, significantly outperforming the current SOTA model DragNUWA (34.5 vs. 36.9). For the evaluation of motion control performance, measured by calculating the Euclidean distance (ObjMC) between the predicted and ground truth trajectories, our model achieved state-of-the-art performance with a score of 302.7 compared to DragNUWA's 326.5. Additionally, in assessing temporal consistency (FVD), which compares the feature distribution between generated and real videos, our model also demonstrated superior temporal coherence, with a score of 510.8, significantly improving by 10.9 over DragNUWA.

\begin{figure*}[t]
\centering
\includegraphics[width=0.95\linewidth]{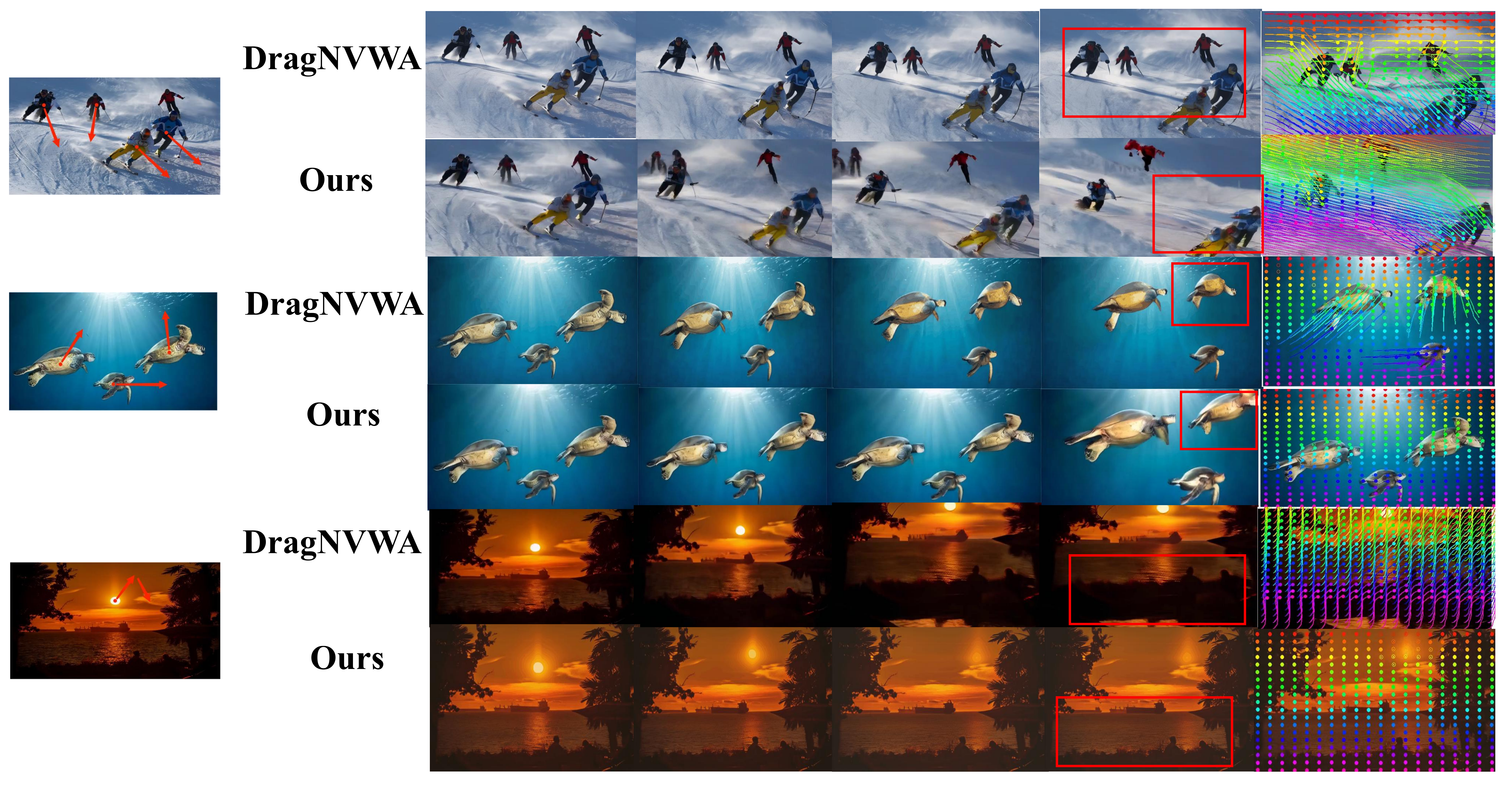}
\vspace{-0.5cm}
\caption{ Visual comparison with DragNUWA. DragNUWA results in appearance distortion (third row), incorrect movement direction (first row), and incorrect camera movement (fifth row), whereas DragEntity can precisely control movement.}
\label{fig7}
\end{figure*}

At the same time, we also conducted visual comparisons on the TedTalk and VIPSeg datasets, as shown in Figure \ref{fig6} and Figure \ref{fig7}. In Figure \ref{fig6}, we can observe that when DragNVWA uses multiple trajectories to control the motion of characters, it results in distortion (third and seventh rows), artifacts (fifth row), and multiple hands (first row). Similarly, As shown in Figure \ref{fig7}, when DragNVWA controls multiple objects, there are issues such as appearance distortion (third row), incorrect movement directions (first row), and incorrect camera movements (fifth row), but our model DragEntity can precisely control motion. Especially when the rising and falling trajectories are applied to the sun, our model can accurately simulate the gradual disappearance of the sun, which aligns with basic physical principles.

\vspace{0.1cm}
\begin{table}[!h]
\centering
\caption{ Performance Comparison on the WebVid Dataset.}
\begin{tabular}{l|ccc}
\hline
Method & ObjMC$\downarrow$ & FVD$\downarrow$ & FID$\downarrow$ \\
\hline
MotionCtrl & 350.6 & 584.2 & 41.8  \\
DragNUWA & 326.5 & 521.7 &  36.9  \\
Ours & \textbf{302.7} & \textbf{510.8} &  \textbf{34.5} \\
\hline
\end{tabular}
\label{tab1}
\end{table}
\vspace{1mm}

\textbf{Ablation Studies.} The entity representation that includes relative spatial relationships is a core component of our work. We keep other conditions constant and only alter the corresponding condition embedding features. Table \ref{tab2} presents the ablation studies for these two aspects. To examine the impact of entity representation, we evaluate performance changes by including or excluding this representation in the final embedding. Since the entity representation mainly influences object motion in generated videos, we focus on comparing ObjMC, whereas FVD and FID metrics assess temporal consistency and overall video quality. With the inclusion of entity representation, the model’s ObjMC significantly improved, reaching a value of 311.4.

\begin{table}[!h]
\centering
\caption{Ablation Study for Entity representation and Spatial Relationship.}
\label{tab:metrics}
\begin{tabular}{lc|ccc}
\hline
 Entity Rep. & Position  & ObjMC$\downarrow$ & FVD$\downarrow$ & FID$\downarrow$ \\
\hline
  & & 368.7 & 563.3 & 42.2 \\
\checkmark &  & 311.4 & 527.5 & 36.1 \\
\checkmark & \checkmark  & 302.7 & 510.8 & 34.5 \\
\hline
\end{tabular}
\label{tab2}
\end{table}

\begin{figure}[t]
\centering
\includegraphics[width=1.0\linewidth]{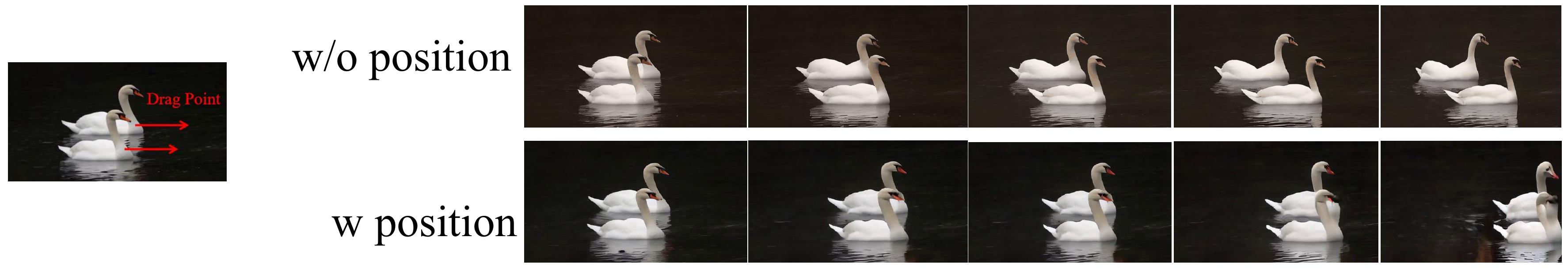}
\caption{Comparison of generated videos with and without position relationships.}
\label{fig8}
\end{figure}

Similar to entity representation, we determine the effectiveness of including spatial positional relationships in the entity representation by observing changes in the ObjMC metric. The spatial relationships between objects led to a performance improvement of 8.7, reaching 302.7. Overall, the highest performance occurs when both entity representation and spatial position relationships are used together. This phenomenon indicates that these two representations have a mutually reinforcing effect, contributing to the precise control of the trajectory. As shown in Figure \ref{fig8}, when the relative positional relationships between objects are not incorporated into our entity representation, it is challenging to maintain the parallel positioning of the two swans. However, when positional relationships are integrated into the entity representation, the relative positions of the two swans remain consistent during movement, demonstrating the necessity of considering positional relationships.

Additionally, we investigated the effect of the loss mask. Table \ref{tab3} details the ablation study concerning the loss mask. In scenarios where the loss mask is not applied, we optimize the MSE loss across every pixel of the entire image. The introduction of the loss mask demonstrates noticeable advantages, leading to an improvement in ObjMC by approximately 13.4.

\begin{table}[!h]
\centering
\caption{ Ablation Study for
Loss Mask.}
\label{tab:metrics}
\begin{tabular}{l|ccc}
\hline
 Loss Mask $M$ & ObjMC$\downarrow$ & FVD$\downarrow$ & FID$\downarrow$ \\
\hline
  & 316.1  & 516.2 & 36.3\\
\checkmark  & 302.7  & 510.8 & 34.5\\
\hline
\end{tabular}
\label{tab3}
\end{table}

\textbf{User Evaluation}
We conducted a user survey to assess video authenticity from MotionCtrl, DragNVWA, and our method, each at 576×320 resolution. Ten volunteers chose the best method based on: 1) structural integrity; 2) trajectory consistency; 3) overall quality. Table. \ref{tab4} shows our method outperforms others in detail, motion, and quality.

\begin{table}[!h]
\centering
\caption{User study results. The percentages indicate the proportion of 10 volunteers who selected the best result from three methods, evaluated from three different perspectives.}
\begin{tabular}{l|ccc}
\hline
Evaluation Criteria & MotionCtrl & DragNVWA & Ours \\
\hline
Structural integrity & 10\% & 25\% & \textbf{65\%} \\
Trajectory consistency & 20\% & 35\% & \textbf{45\%}  \\
Overall feeling & 10\% & 30\%  & \textbf{60\%} \\
\hline
\end{tabular}
\label{tab4}
\end{table}

\section{Conclusion}
In this paper, we introduce a new trajectory-based motion control method and present two new insights: 1) Pixel points controlled by trajectories do not adequately represent entities. 2) When multiple trajectories act on multiple objects, the objects can maintain relative spatial relationships to preserve spatial consistency. To address these two challenges, we propose DragEntity, which utilizes latent features to represent each entity. Our proposed entity representation incorporates relative spatial position relationships as a self-domain embedding, enabling the control of entity motion in the image while maintaining structural integrity. Experiments validate the superiority of our method over existing approaches, demonstrating its ability to effectively generate fine-grained videos.



\bibliographystyle{ACM-Reference-Format}
\bibliography{sample-sigconf}


\end{document}